\documentclass[preprint,12pt,authoryear]{elsarticle}

\usepackage[T1]{fontenc}    
\usepackage{hyperref}  
\usepackage{amssymb}
\usepackage{url}  

\usepackage{booktabs}       
\usepackage{amsfonts}       
\usepackage{nicefrac}       
\usepackage{microtype}      
\usepackage{color}
\usepackage{microtype}
\usepackage{graphicx}
\usepackage{subfigure}
\usepackage{amsmath}
\usepackage{booktabs} 
\usepackage{bm}
\usepackage{multirow}
\usepackage{float}
\usepackage{lineno}
\usepackage{quotes}

\usepackage{adjustbox}

\newcommand{\re}[1]{\textcolor{black}{#1}}

\begin{document}




\title{Detecting Out-of-distribution Samples via Variational Auto-encoder with Reliable Uncertainty Estimation}


\author[1,5]{Xuming Ran\fnref{fn1},\corref{cor1}}
\author[2]{Mingkun Xu\fnref{fn1}}
\author[3]{Lingrui Mei}
\author[4,6]{Qi Xu}
\author[1]{Quanying Liu\corref{cor1}}

\fntext[cor1]{Equal contribution: Xuming Ran and Mingkun Xu}
\cortext[cor1]{Corresponding author: Xuming Ran (ranxuming@gmail.com) and Quanying Liu (liuqy@sustech.edu.cn)}

\address[1]{Shenzhen Key Laboratory of Smart Healthcare Engineering, Department of Biomedical Engineering, Southern University of Science and Technology, Shenzhen 518055, China}
\address[2]{Center for Brain Inspired Computing Research, Department of Precision Instrument, Tsinghua  University, Beijing 100084, China}
\address[3]{China Automotive Engineering Research Institute, Chongqing 401122, China}
\address[4]{School of Artifical Intelligence, Electronic and Electrical Engineering, School of Artifical Intelligence Dalian University of Technology, Dalian 116024, China}
\address[5]{College of Mathematics and Statistics, Chongqing Jiaotong
University, Chongqing 400074, China}
\address[6]{College of Computer Science and Technology, Zhejiang University, Hangzhou 310027, China}

\begin{abstract}
Variational autoencoders (VAEs) are influential generative models with rich representation capabilities from the deep neural network architecture and Bayesian method. 
However, VAE models have a weakness that assign a higher likelihood to out-of-distribution (OOD) inputs than in-distribution (ID) inputs. 
To address this problem, a reliable uncertainty estimation is considered to be critical for in-depth understanding of OOD inputs. 
In this study, we propose an improved noise contrastive prior (INCP) to be able to integrate into the encoder of VAEs, called INCPVAE. 
\re{INCP is scalable, trainable and compatible with VAEs, and it also adopts the merits from the INCP for uncertainty estimation.}
Experiments on various datasets demonstrate that compared to the standard VAEs, our model is superior in uncertainty estimation for the OOD data and is robust in anomaly detection tasks.
\re{The INCPVAE model obtains reliable uncertainty estimation for OOD inputs and solves the OOD problem in VAE models.}

\end{abstract}

\maketitle
\section{Introduction}
\re{The out-of-distribution (OOD) data has a significantly different distribution from the training in-distribution (ID) data. To make reliable and safe decisions, the deep learning models in real-world applications require to identify whether the testing data is the OOD data.} 
Likelihood models are considered to naturally own the ideal capability of detecting OOD inputs, due to the intuitive assumption that these models assign lower likelihoods to the OOD inputs than the in-distribution (ID) inputs~\cite{bishop1994novelty}. However, previous works have reported that some deep generative models, such as variational auto-encoders (VAEs)~\cite{kingma2013auto,rezende2014stochastic}, Pixel CNN~\cite{van2016conditional} and Glow~\cite{kingma2018glow}, all based on likelihood models, are not able to correctly detect OOD inputs~\cite{nalisnick2018deep,hendrycks2018deep,choi2018waic,lee2017training,nalisnick2019detecting,maaloe2019biva}. Counter-intuitively, the OOD inputs are assigned higher likelihoods than the ID inputs, which is not in line with the assumption. Hence, when we employ the likelihood model as a detector on OOD detection tasks or general generation tasks, it is necessary to ensure that the adopted model possesses a good understanding and performance for OOD inputs. 

\re{The phenomenon that VAE models assign higher likelihoods to OOD inputs than ID inputs is called the OOD problem, and it was first reported by~\cite{nalisnick2018deep} in 2018. Since then, it has been an increasingly popular topic in the field of generative models.
Some studies have made great efforts to explain the reasons for this empirical phenomenon~\cite{nalisnick2019detecting,serra2019input,butepage2019modeling}. 
For instance, Bütepage et al. demonstrates that it is caused by model assumptions and evaluation schemes, where the oversimplified likelihood function (e.g., iid Bernoulli or iid Gaussian) assumed in the VAE model affects the judgment of the data distribution of the ID inputs~\cite{butepage2019modeling}. However, the true likelihood function is often unknown and more complicated, which has certain deviations from the assumed one. 
In some datasets, local evaluations with the approximated posterior can lead to overconfidence. Nalisnick et al. conjectures that the high-likelihood region conflicts with the typical set of the model~\cite{nalisnick2019detecting}. Serrà et al. posits that the complexity of the input data will have a strong impact on likelihood-based models~\cite{serra2019input}. }

Many approaches have been studied to solve the OOD detection problem in generative models. Some studies have suggested that likelihood models with reliable uncertainty estimates may help improve OOD detection~\cite{nalisnick2018deep,choi2018waic}. 
In addition, noise contrastive priors (NCPs) are a specific prior in the data space for neural networks, encouraging network weights to not only explain the ID inputs, but also capture the high uncertainty of OOD samples~\cite{hafner2018reliable}. Thus, NCPs might help the uncertainty estimates of the OOD data.
Inspired by these two viewpoints, we propose a novel method, named Improved Noise Contrastive Priors Variational Auto-encoder (INCPVAE), to allow VAE models to obtain reliable uncertainty estimates thereby solving the OOD detection problem. 
\re{Although the original NCPs are often applied to classifier models, they cannot be directly applied to the VAE framework. Therefore, we have to improve the loss function of NCP (called the improved NCP, INCP) to make it suitable for the VAE framework. }
The INCP is integrated into the encoder of VAE, so that OOD samples can be generated by adding Gaussian noise to the origin ID inputs. 
Since using the simple likelihood function of VAE often leads to poor performance on OOD detection tasks, we exploit the INCP-KL divergence of INCPVAE, rather than the likelihood, for detecting OOD inputs. 
Our experiments show that compared to the traditional VAEs, our INCPVAE can reduce the overconfidence when facing OOD data and obtain better performances of OOD detection. The main contributions of this paper are as follows:

\begin{itemize} 

\item We propose an improved noise contrastive prior to fit the VAE framework (\textbf{Sec~\ref{incpvae}}). To the best of our knowledge, this is the first work to use the noise contrastive prior to obtain reliable uncertainty estimates in unsupervised generative models. 

\item We present a tailored metric (the ELBO Ratio) in the INCPVAE framework to estimate the uncertainty (\textbf{Sec~\ref{mue}}), which can achieve reliable uncertainty estimation and enhanced robustness \re{(\textbf{Sec~\ref{uer}})}.

\item We propose a novel OOD detection method by using the INCP-KL ratio of INCPVAE (\textbf{Sec~\ref{incpkl}}). Through a number of experiments on the challenging OOD cases, we demonstrate that INCPVAE can learn the true characterization of OOD inputs, and achieves state-of-the-art (SOTA) performance in OOD detection \re{(\textbf{Sec~\ref{ood}})}.

\end{itemize}

\section{Related Work}

\textbf{OOD detection:} There are many neural network tools that can be used to perform pattern recognition, image classification, and OOD detection tasks, such as spike neural networks~\citep{ciagnsupervise,liu2020uns,xu2018cs} and convolutional neural networks~\citep{lee2018simple, xu2019Overfitti}.
The OOD detection permits a system to reject a novel input rather than assigning it an incorrect label; therefore the ability to detect OOD data is essential for machine learning models.
From the algorithm perspective, there are two categories of mainstream approaches for OOD detection, i) the supervised/discriminative approaches and ii) the unsupervised/generative approaches~\cite{daxberger2019bayesian}.
\re{ Most existing methods belong to the supervised model. For example, the classifiers are trained by both the OOD data and ID data to learn a decision boundary between ID and OOD inputs, which can be used for OOD detection. Liang et al. present an OOD detector with neural networks (called ODIN) which uses softmax function to maximize the difference between likelihoods of ID data and OOD data, while the model parameters are tailored to each OOD source~\cite{liang2017enhancing}. Lakshminarayanan et al. propose an ensemble method for OOD detection, which independently trains multiple models with random initializations of network parameters and randomly shuffled training inputs ~\cite{lakshminarayanan2017simple}.  
Some previous studies show that these supervised methods can to some extent prevent the poorly-calibrated neural networks from incorrectly high-confidence on OOD inputs~\cite{liang2017enhancing, lakshminarayanan2017simple, devries2018learning}. This capability can be used in various applications, including anomaly detection~\cite{hendrycks2016baseline,vyas2018out,PidhorskyiGenerative} and adversarial defense~\cite{song2018constructing}.}  However, these methods can only be applied to task-dependent scenarios. This is a severe limitation, for the anomalous data in real-world applications rarely knows in advance.

In contrast, the unsupervised approaches aim to solve the OOD detection problem by training deep generative models in a more general manner, among which density estimation is widely applied~\cite{kingma2018glow,oord2016pixel}. \re{For example, Choi et al. use generative model with Watanabe-Akaike information criterion (WAIC) for detecting OOD~\cite{choi2018waic}. Although this work performs well in practice, it does not explicitly solve the problem of typicality~\cite{choi2018waic,nalisnick2019detecting}. Denouden et al. propose a method that incorporates both reconstruction loss and the Mahalanobis distance~\cite{lee2018simple} in the latent space as an OOD detection score~\cite{denouden2018improving}. Ren et al. propose a likelihood ratio method for deep generative models to detect the OOD data~\cite{ren2019likelihood}. Zhang et al. studied the intrinsic robustness of typical image distributions by using conditional generative models~\cite{zhang2020understanding}. They proved a fundamental bound on the intrinsic robustness, that is, the underlying data distribution can be captured by a conditional generative adversarial network.} 
However, as mentioned, the likelihood estimation in deep generative models are not reliable for OOD detection. Many studies have attempted to explain the reasons and seek the solutions~\cite{nalisnick2019detecting,serra2019input,butepage2019modeling}. So far, an efficient and robust solution for OOD detection is still missing and urgently needed.

\textbf{Uncertainty estimation:} Uncertainty estimation is highly associated with OOD detection. The goal of uncertainty estimation is to generate a calibrated confidence measure for the predicted distribution which can be used in the OOD detection. 
The uncertainty estimation in MC Dropout~\cite{gal2016dropout}, Deep-Ensemble~\cite{lakshminarayanan2017simple} and ODIN~\cite{liang2017enhancing} involves presenting a calibrated
predictive distribution by classifiers. 
Alternatively, variational information bottleneck (VIB) conducts OOD detection via divergence estimation in latent space~\cite{alemi2018uncertainty}. However, these existing methods are model-dependent and rely heavily on task-specific information to obtain a comprehensive estimate of uncertainty. Therefore, a more general and task-independent method is of high needs. 

Recent studies have suggested that likelihood models with reliable uncertainty estimation can help to mitigate the high OOD likelihood problem for generative models in a task-independent manner~\cite{nalisnick2018deep,choi2018waic}. \re{For example, Meronen et al. studied the influence of neural network activation functions and the Matérn family of kernels on the uncertainty estimation~\cite{meronen2020stationary}. }
Moreover, as an influential and generally-used class of likelihood-based generative models in unsupervised learning, VAEs may be a good OOD detector. It assumes that the model assigns higher likelihoods to the samples from the ID data than the OOD data. \re{ NCPs can inject variability or insensitivity into the model especially into regions that do not exhibit that otherwise after training. In this sense, NCPs can be considered as a part of model specification to get better estimation of uncertainty and therefore help model inference.}
In this study, we provide a novel hybrid framework that bridges NCPs with VAEs and generates OOD data by adding Gaussian noise, to help both the reliability of uncertainty estimation and model independence in OOD detection.  

\section{Method}

\subsection{Improved Noise Contrastive Priors}
\label{NCP}

NCPs has been proposed to obtain reliable uncertainty estimates by employing an input prior to the ID inputs $\bm{x}$ and OOD inputs $\bm{\tilde{x}}$ and an output prior which is a wide distribution given these inputs~\cite{hafner2018reliable}. However, NCPs are not suitable for VAE framework. 
In this work, we modify the loss function to make the original NCPs fit the VAE framework, to obtain uncertainty through the VAE model. 
We add Gaussian noise to ID images to generate OOD data.

\begin{figure}[H]
\centering
\includegraphics[width=0.6\textwidth]{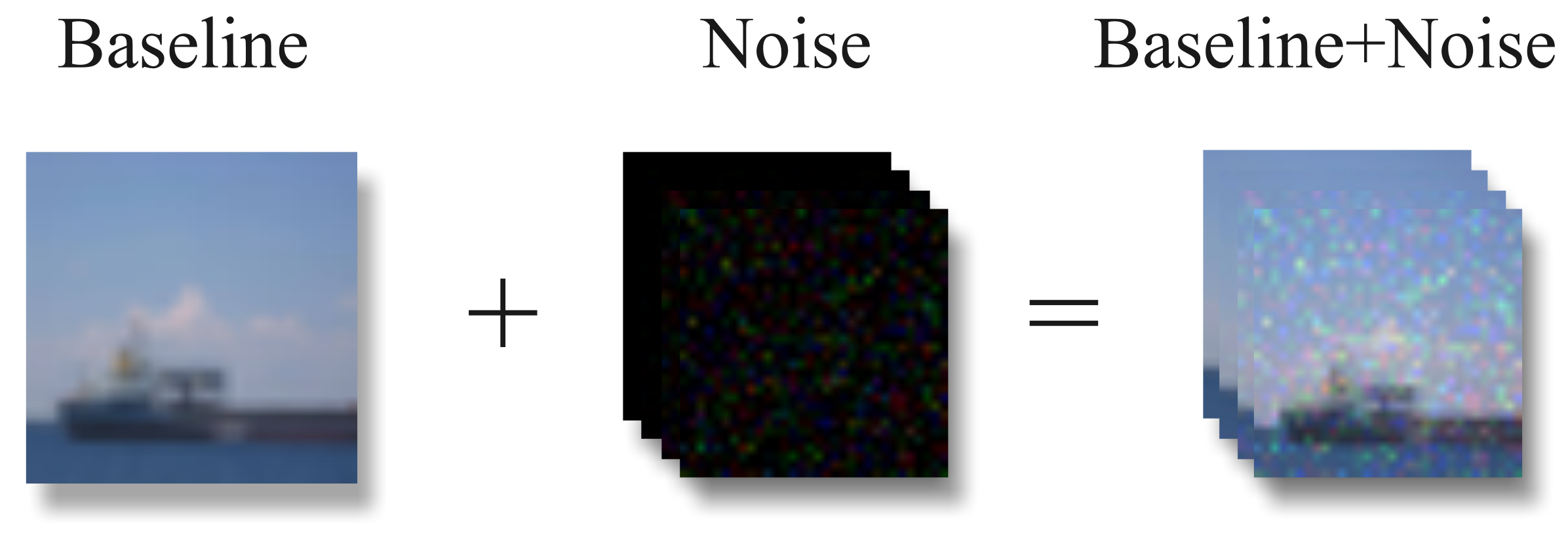}
\caption[]{Generating OOD samples by adding Gaussian Noise to the baseline data. The baseline data is sample from the original image dataset (e.g., FashionMNIST, MNIST, CIFAR10, SVHN). We add the Gaussian Noise at three levels to generate the OOD sample with different complexity. The Baseline+Noise is the generated OOD sample.}
\label{figood}
\end{figure}

\textbf{Generating OOD Inputs:} OOD samples can be generated by sampling from the distribution boundary of the ID data with high uncertainty~\cite{lee2017training}.
Inspired by noise contrastive estimation~\cite{gutmann2010noise,mnih2013learning}, Hafner et al.~\cite{hafner2018reliable} proposed a NCP-based algorithm, where a complement distribution is approximated by random noise. To obtain OOD inputs $\bm{\tilde{x}}$, we add Gaussian noise $\bm{\epsilon}$ into the continuous ID inputs $\bm{x}$, formulated as $\bm{\tilde{x}}= \bm{x} + \bm{\epsilon} $ (\textbf{See Fig~\ref{figood}}). 
The marginal distribution of OOD inputs ${p}_{{o}}(\bm{\tilde{x}})$ is derived in Eq.(\ref{e1}) as follows.

\begin{equation}
\label{e1}
\begin{aligned} 
 {p}_{{o}}(\bm{\tilde{x}})  =\int_{\bm{x}} {p}_{{i}}(\bm{x}) \mathcal{N}\left(\tilde{\bm{x}}-\bm{x} \mid \mu_{}, \sigma_{}^{2}\bm{I}\right) d \bm{x},
\end{aligned} 
\end{equation}
where ${p}_{{i}}(\bm{{x}})$ denotes the distribution density of ID inputs; $\mu_{}$ and $\sigma_{}^{2}$ are the mean and variance of Gaussian noise, respectively. In order to make the noise contrastive prior homogeneous in all directions of the data manifold, we set $\mu_{}=0$. The variance $\sigma_{}^{2}$ is a hyperparameter to tune the sampling distance from the boundary of the training ID distribution. The higher the variance $\sigma_{}^{2}$, the higher the complexity of OOD inputs.

\textbf{Data Priors:} The data priors consist of an input prior $p_{ {}}(\bm{x})$ and an output prior $p_{ {}}(\bm{z}|\bm{x})$. 
To obtain a reliable uncertainty estimation by the VAE model, appropriate input priors (including a prior on OOD inputs) should be set. 
A good output prior should be a high-entropy distribution, which serves as the high uncertainty of the VAE's target output for a given OOD input. 
The data priors in our model are listed as follows:

\begin{equation}
\label{e2}
\begin{aligned} 
\text{OOD input prior: } & {\tilde p}(\bm{\tilde{x}}) = p_{o}(\bm{\tilde x})
\\
\text{OOD output prior: } & \re{{\tilde p}_{ { }}({\bm{\tilde{z}}} \mid \bm{\tilde{x}}) =\mathcal{N}\left(  \bm{\tilde{z}} \mid \mu_{\bm{\tilde x}}, \sigma_{\bm{\tilde x}}^{2}\bm{I}\right)},
\end{aligned} 
\end{equation} 
where $p_{o}(\bm{\tilde x})$ is the prior distribution of OOD inputs; $\mu_{\bm{\tilde x}} $ and $\sigma_{\bm{\tilde x}}^{2}$ are the hyperparameters of OOD output priors to tune the mean and the uncertainty in the target outputs. 

\textbf{Loss Function:} \re{KL divergence is not symmetric, and it has a forward version and a reverse 
version~\cite{zhang2019variational}. In the original NCPs~\cite{hafner2018reliable}, both the difference metrics between the distribution $ p_{ {}}(\bm {z} \mid \bm{x})$ and $q_{{\theta}}(\bm {z} \mid \bm{x}) $ and between $ {\tilde p}_{ { }}(\bm {\tilde z} \mid \bm{\tilde x})$ and $q_{ {\theta  }}(\bm {\tilde z} \mid \bm{\tilde x} ) $ adopt the forward KL divergence. However, the VAE uses the reverse KL divergence as its basic metric in loss function, which leads to the inconsistency in optimization strategy and direction. Therefore, this poses an intractable challenge for constructing a unified optimization framework by incorporating NCP organically, where the forward KL divergence is not compatible for VAE. To better tackle the challenge and incorporate the NCP into the VAE framework, we proposed the improved NCP (INCP) method by integrating the reverse KL divergence into the NCP. To train INCPs, we modify the loss function as follows:
\begin{align}
\label{e3}
\begin{split}
 \mathcal{L}(\theta) &=\bm{E}_{q_{ {\theta}}(\bm {z} \mid \bm{x})}\big[\bm{D}_{{KL}}\left[q_{{\theta}}(\bm {z} \mid \bm{x}) \mid \mid p_{ {}}(\bm {z} \mid \bm{x})\right]\big] \\
  &+\gamma \bm{E}_{q_{ { \theta }}(\bm { \tilde z} \mid \bm{ \tilde x})}\big[\bm{D}_{{KL}}\left[q_{ {\theta  }}(\bm {\tilde z} \mid \bm{\tilde x}) \mid \mid {\tilde p}_{ { }}(\bm {\tilde z} \mid \bm{\tilde x})\right]\big], 
\end{split}
\end{align}
where  ${\tilde p}(\bm {\tilde z} \mid \bm{\tilde x})$ denotes OOD data priors, $\theta$ is the parameter of neural network. A hyper-parameter $\gamma$ denotes the trade-off between the ID and OOD output priors. INCPs can be trained by minimizing this loss.
Notice that in the Eq.~\ref{e3}, by minimizing the reverse KL divergence in the first term, the neural network is trained to suit for the true ID data outputs prior. And an analogous term on the OOD data outputs prior is added in the second term. This loss function simultaneously optimizes the ID and OOD outputs prior for two distinct targets (\textit{i.e.}, the true ID data outputs prior $\&$ the assumed OOD data outputs prior). In contrast, the origin NCP loss~\cite{hafner2018reliable} hardly integrates the ID and OOD conditional distribution into one target in the VAE framework. }

\subsection{Variational Autoencoder}

VAEs~\cite{kingma2013auto,rezende2014stochastic} are a class of latent variable models optimized by the maximum marginal likelihood of an observation variable. The marginal likelihood $ p(\bm{x}) $ can be written as follows:

\begin{equation}
\label{e4}
\begin{aligned} 
\log p(\bm{x})= & \bm{E}_{\bm{z} \sim q_{\theta}(\bm{z} \mid \bm{x})}[\log p_{\phi}(\bm{x} \mid \bm{z})]-\bm{D}_{KL}[q_{\theta}(\bm{z} \mid \bm{x}) \| p(\bm{z})]\\
&+ \bm{D}_{KL}[q_{\theta}(\bm{z} \mid \bm{x}) \| p(\bm{z} \mid \bm{x})],
\end{aligned}
\end{equation}
where $p(\bm z)$ and $p(\bm{z} \mid \bm{x})$ are the ID input/output priors (e.g., Vamp Prior~\cite{tomczak2017vae}, Resampled Prior ~\cite{bauer2018resampled}). \re{In this study, $p(\bm z)$ is instantiated by a standard normal distribution, and $p(\bm{z} \mid \bm{x})$ is the true posterior distribution corresponding to $p(\bm z)$.
The encoder $q_{\theta}(\bm{z} \mid \bm{x})$ and the decoder $p_{\phi}(\bm{x} \mid \bm{z})$ are modeled by two neural networks parameterized with $\theta$, $\phi$, respectively.}
Specifically, $q_{\theta}(\bm{z} \mid \bm{x})$ represents the variational posterior (the encoder) which is implemented by a Guassian distribution, and $p_{\phi}(\bm{x} \mid \bm{z})$ is the generative model (the decoder) which is implemented by a Bernoulli distribution. 

However, the true posterior $p(\bm{z} \mid \bm{x})$ cannot be computed analytically. Assuming that the variational posterior $q_{\theta}(\bm{z} \mid \bm{x})$  has a arbitrarily high-capacity for modeling, $q_{\theta}(\bm{z} \mid \bm{x})$ can learn to approximate the intractable $p(\bm{z} \mid \bm{x})$ and the reverse KL divergence between $q_{\theta}(\bm{z} \mid \bm{x})$ and  $p(\bm{z} \mid \bm{x})$ goes to zero. Thus, we train the VAE with ID samples, or OOD samples, to maximize the following objective variational evidence lower bound, which are called ELBO$_I$ for ID samples, and ELBO$_O$ for OOD samples.

\begin{equation}
\label{e5}
\begin{aligned} 
\text{ELBO}_I(\phi, \theta)=  \bm{E}_{\bm{z} \sim q_{\theta}(\bm{z} \mid \bm{x})}[\log p_{\phi}(\bm{x} \mid \bm{z})]-\bm{D}_{KL}[q_{\theta}(\bm{z} \mid \bm{x}) \| p(\bm z)]
\\
\re{
\text{ELBO}_O(\phi, \theta)=  \bm{E}_{\bm{\tilde z} \sim q_{\theta}(\bm{ \tilde z} \mid \bm{\tilde x})}[\log p_{\phi}( \bm{\tilde x} \mid \bm{ \tilde z})]-\bm{D}_{KL}[q_{\theta}(\bm{ \tilde z} \mid \bm{\tilde x}) \| {\tilde p}(\bm{ \tilde z})]}
\end{aligned}
\end{equation}
where $q_{\theta}(\bm{z} \mid \bm{x})$  and $q_{\theta}(\bm{\tilde z} \mid \bm{\tilde x})$ are the variational posteriors which approximate the true posteriors (\textit{i.e.,} $p_{}(\bm{z} \mid \bm{x})$  and \re{ ${\tilde p}(\bm{\tilde z} \mid \bm{\tilde x})$}), given the ID input $ \bm{\tilde x}$ and the OOD input $\bm{x}$, respectively.
For a given dataset, the marginal likelihood $ p(\bm{x}) $ is a constant. Substituting Eq.~\ref{e5} to Eq.~\ref{e4}, we obtain

\begin{equation}
\label{e6}
\begin{aligned} 
\log p(\bm{x})= \text{ELBO}_I(\phi, \theta) + \bm{D}_{KL}[q_{\theta}(\bm{z} \mid \bm{x}) \| p(\bm{z} \mid \bm{x})]  = \text{const.}
\end{aligned}
\end{equation}

From Eq.~\ref{e6}, it is obvious that maximizing ELBO$_I$ is equivalent to  minimizing the KL-divergence between $q_{\theta}(\bm{z} \mid \bm{x})$ and  $p(\bm{z} \mid \bm{x})$. \re{Likewise, maximizing ELBO$_O$ is equivalent to minimizing the reverse KL divergence between $q_{\theta}(\bm{ \tilde z}  \mid \bm{\tilde x})$  and ${\tilde p}_{\theta}(\bm{ \tilde z}  \mid \bm{\tilde x})$.}

\subsection{INCP Variational Autoencoder}
\label{incpvae}
INCPVAE consists of an encoder and a decoder, and the INCPs are imposed on the encoder network of VAE. The INCPVAE is trained on both ID and OOD inputs by minimizing ELBO$_I$ and ELBO$_O$ as shown in Eq.~\ref{e5}. We \re{define the total ELBO of INCPVAE, $\text{ELBO}_{INCP}(\phi, \theta)$, as follows,}

\begin{equation}
\label{eq19}
\text{ELBO}_{INCP}(\phi, \theta)=\text{ELBO}_I(\phi, \theta)+ \gamma \text{ELBO}_O(\phi, \theta),
\end{equation}
\re{where the hyper-parameter $\gamma $ is a setting as a trade-off between ELBO$_I$ and ELBO$_O$. }

We assume the variational posterior $q_{\theta}(\bm{z} \mid \bm{x})$ for ID inputs has high-capacity for modelling, then true posterior $p(\bm{z} \mid \bm{x})$ can be approximated by $q_{\theta}(\bm{z} \mid \bm{x})$. 
Since the OOD outputs prior ${\tilde p}_{ { }}({\bm{\tilde{z}}} \mid \bm{\tilde{x}})$ is defined in Eq.~\ref{e2}, the true OOD data posterior \re{ ${\tilde p}_{}(\bm{\tilde z} \mid \bm{\tilde x})$} is:

\begin{equation}
\begin{aligned} 
 \re{{\tilde p}_{}(\bm{\tilde z} \mid \bm{\tilde x})}= \mathcal{N}\left(  \bm{\tilde z} \mid \mu_{\bm{\tilde x}}, \sigma_{\bm{\tilde x}}^{2}\bm{I}\right),
\end{aligned} 
\end{equation}
where $\mu_{\bm{\tilde x}} = \mu_{\bm{x}}$ and $(\mu_{\bm{x}} \sim q_{\theta}(\bm{z} \mid \bm{x}))$; $\sigma_{\bm{\tilde x}}^{2}$ is a hyper-parameter to tune the uncertainty in the outputs. The higher  $\sigma_{\bm{\tilde x}}^{2}$ , the higher the output uncertainty. 
The reverse KL divergence between $q_{\theta}(\bm{\tilde z} \mid \bm{\tilde x}) $ and \re{$ {\tilde p}(\bm{\tilde z} \mid \bm{\tilde x}) $} (called INCP-KL) becomes tractable and can be analytically computed. 
\re{ From Eq.~\ref{eq19} }, maximizing the ELBO of INCPVAE can be replaced by minimizing the following loss function:

\begin{equation}
\label{eq9}
\mathcal{L}_{INCPVAE}(\phi, \theta) = -\text{ELBO}_I(\phi, \theta) +\gamma \underbrace{ \mathop{ {\bm{D}_{KL}[q_{\theta}(\bm{\tilde z}| \bm{\tilde x}) \| \ \re{{\tilde p}(\bm{\tilde z}| \bm{\tilde x})}]}} }\limits_{  \quad \mathbf{INCP-KL} \quad \bm{Loss} \quad }
\end{equation}

\re{Notably, the first term in Eq.~\ref{eq9} minimizes the negative ELBO$_I$, which is equivalent to maximizing ELBO$_I$. The second term in Eq.~\ref{eq9} minimizes INCP-KL for OOD data, which is equivalent to maximizing ELBO$_O$, according to Eq.~\ref{e6}. In this study, we set the hyperparameter $\gamma=1$.}

\subsection{Metrics for Uncertainty Estimation: ELBO Ratio}
\label{mue}

We proposed the objective variational evidence lower bound ratio (ELBO Ratio) for an uncertainty estimation metric of VAE. According to Eq.~\ref{e5}, we compute the ELBO of each ID sample and find the maximum one (called $\textbf{ELBO}_I(\bm{x_{max}})$ ). The ELBO Ratio for input data $\bm {x_{0}}$, $\mathcal{U}(\bm {x_0})$, is defined as 

\begin{equation}
\label{e}
\begin{aligned}
\mathcal{U}(\bm {x_0}) = \frac{ \textbf{ELBO}(\bm {x_{0}}) }{ \textbf{ELBO}_I(\bm{x_{max}})},
\end{aligned}
\end{equation}
The ELBO ratio $\mathcal{U}(\bm {x_0})$  measures the degree of uncertainty on data $\bm {x_0}$. The greater $\mathcal{U}(\bm {x_0})$ , the higher uncertainty $\bm {x_0}$.

\subsection{OOD Detection based on INCP-KL Ratio}
\label{incpkl}

\textbf{INCP-KL ratio:} The likelihood of VAE has been used for OOD detection. However, it is reported that the OOD inputs have a higher likelihood than ID inputs that occur in some datasets (e.g., FashionMNIST vs MNIST, CIFAR10 vs SVHN). To solve this problem, Likelihood Ratios for OOD detection has been proposed~\cite{ren2019likelihood}. In Eq.~\ref{eq9}, the second term of the INCPVAE loss is \re{the reverse KL divergence between the OOD variational posterior ($q_{\theta}(\bm{\tilde z}| \bm{\tilde x})$) and the true OOD posterior ($ {\tilde p}(\bm{\tilde z}| \bm{\tilde x})$), which is called INCP-KL.} We find that INCP-KL of the OOD test samples (e.g., Baseline+Noise, in \textbf{Fig~\ref{figood}}) is smaller than the ID samples from other distribution. Inspired by it, we use the INCP-KL Ratio for OOD detection. \re{We calculate the INCP-KL divergence for all OOD training samples $\tilde x$ in the OOD dataset, and then find the OOD sample with maximum INCP-KL (called $OOD_{max}$). The INCP-KL Ratio for input data $\bm {x_{0}}$, $KLR(\bm {x_0})$, is defined as
\begin{equation}
\label{eqs}
\begin{aligned}
OOD_{max} & = \mathop{argmax}\limits_{ \tilde x_{}} \mathop{ {\bm{D}_{KL}[q_{\theta}(\bm{\tilde z}| \bm{\tilde x}) \| \ {\tilde p}(\bm{\tilde z}| \bm{\tilde x})}]}
\\
KLR(\bm {x_0}) & = \frac{\bm{D}_{KL}[q_{\theta}(\bm{ z_0}| \bm{x_0}) \| {\tilde p}(\bm{ \tilde z}| \bm{ \tilde x})]} { \bm{D}_{KL}(OOD_{max})}
\end{aligned}
\end{equation}
where $D_{KL}(OOD_{max})$ is INCP-KL divergence of the OOD sample, $OOD_{max}$.}

\re{\textbf{OOD detection criterion:} The OOD detection based on INCP-KL Ratio is as follows: 
\begin{equation}
\label{ood-detection}
\begin{aligned}
\bm{Label}(\bm {x_0}) & =\left\{
\begin{array}{rcl}
0,       &\text{if}      & {{KLR}}(\bm {x_0}) > \alpha\\
1,       &\text{if}     & {{KLR}}(\bm {x_0}) \leq \alpha
\end{array} \right.
\end{aligned}
\end{equation} 
where $\alpha$ is the decision threshold. In our study, we set $\alpha=1$.  $\bm{Label}(\bm {x_0})=1$ represents that the test sample $\bm {x_0}$ is detected as OOD data; $\bm{Label}(\bm {x_0})=0$ represents that $\bm {x_0}$ is detected as ID data. }

\section{Experiments and Results}

\subsection{Experimental settings}
To evaluate our method and compare with other existing methods, we conduct experiments on multiple datasets. There are two tasks involved in the experiments: the uncertainty estimation task and the OOD detection task.

To obtain the ground-truth OOD data, we synthesize OOD data by adding Gaussian noise to the baseline data (ID data), as \textbf{Sec~\ref{NCP}} described. The baseline data are from FashionMNIST, MNIS, CIFAR10 and SVHN. Three levels of Gaussian noise ($\mu=0$, $\sigma = \sigma_0, \sigma_1, \sigma_2$) are generated to represent three levels of uncertainty in OOD data.
The detailed settings of the baseline ID data and the synthesized OOD data are in \textbf{Appendix A}. 

To replicate the OOD phenomenon in VAE models, we conduct the likelihood tests of ID and OOD data, following the experimental settings in~\cite{nalisnick2018deep} (Details are shown in \textbf{Appendix B}). Specifically, we train the traditional VAE on the training set (ID samples) and compute the likelihoods of 1000 random samples from the test set (including both ID samples and their corresponding OOD samples). We exhibit the histogram of the marginal likelihoods of the 1000 tests on VAE (\textbf{Fig~\ref{fig-A2}}). 

\begin{figure}[H]
\centering
\subfigure[FashionMNIST]{\includegraphics[width=0.49\textwidth]{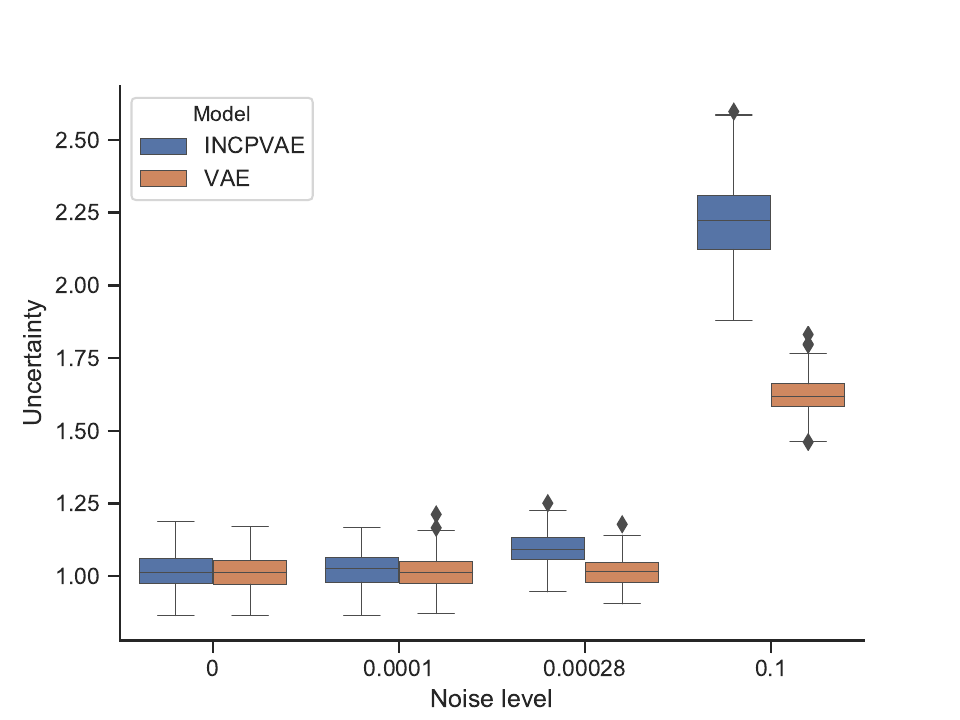}}
\subfigure[MNIST]{\includegraphics[width=0.49\textwidth]{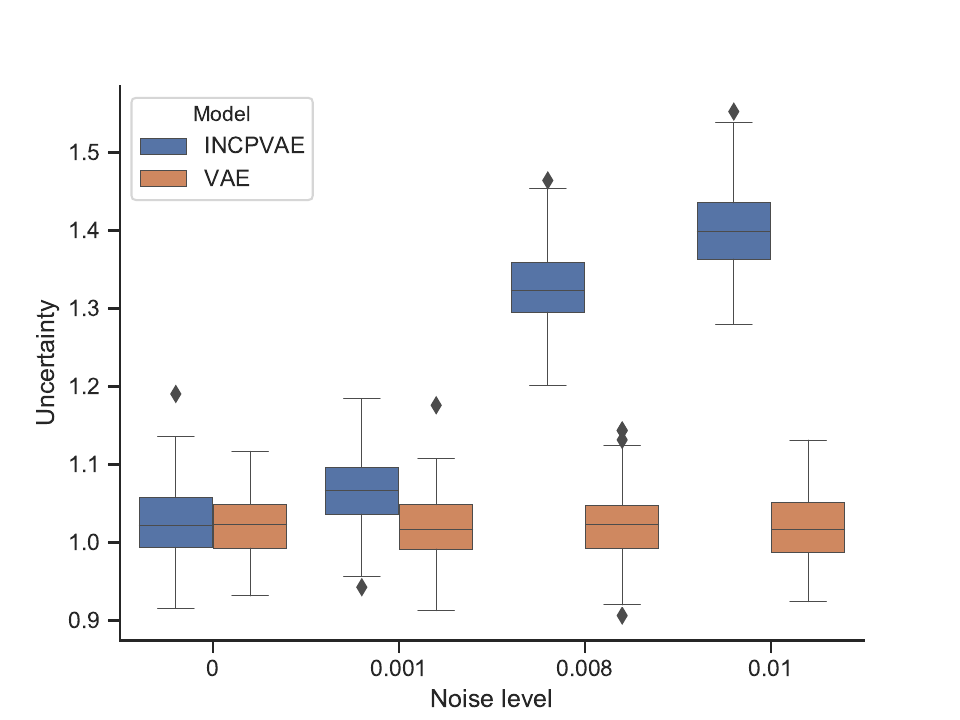}} 

\subfigure[CIFAR10]{\includegraphics[width=0.49\textwidth]{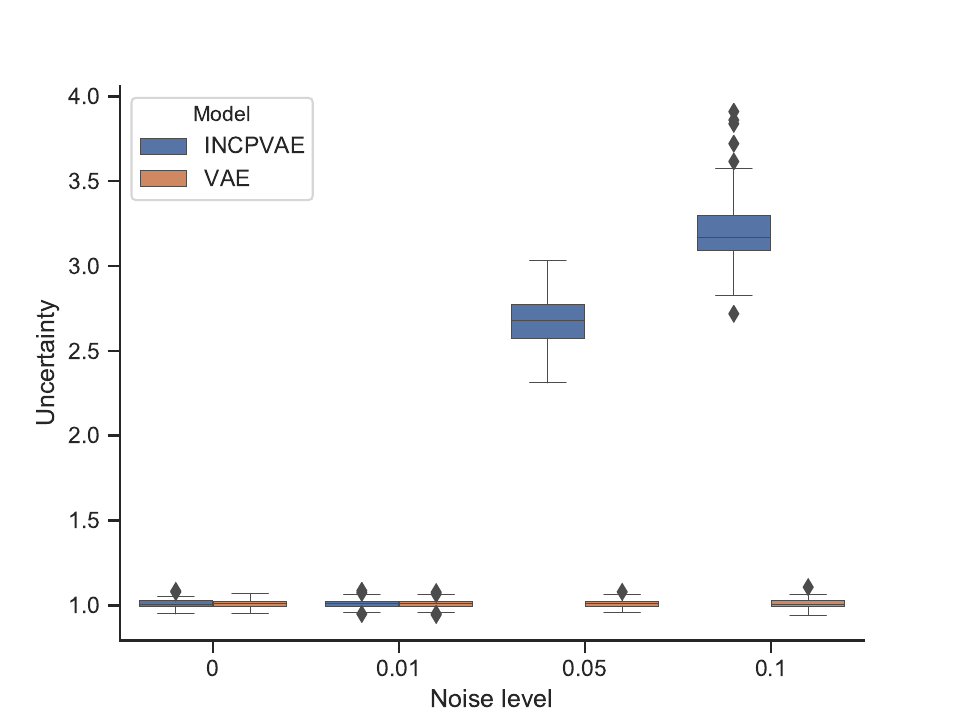}}
\subfigure[SVHN]{\includegraphics[width=0.49\textwidth]{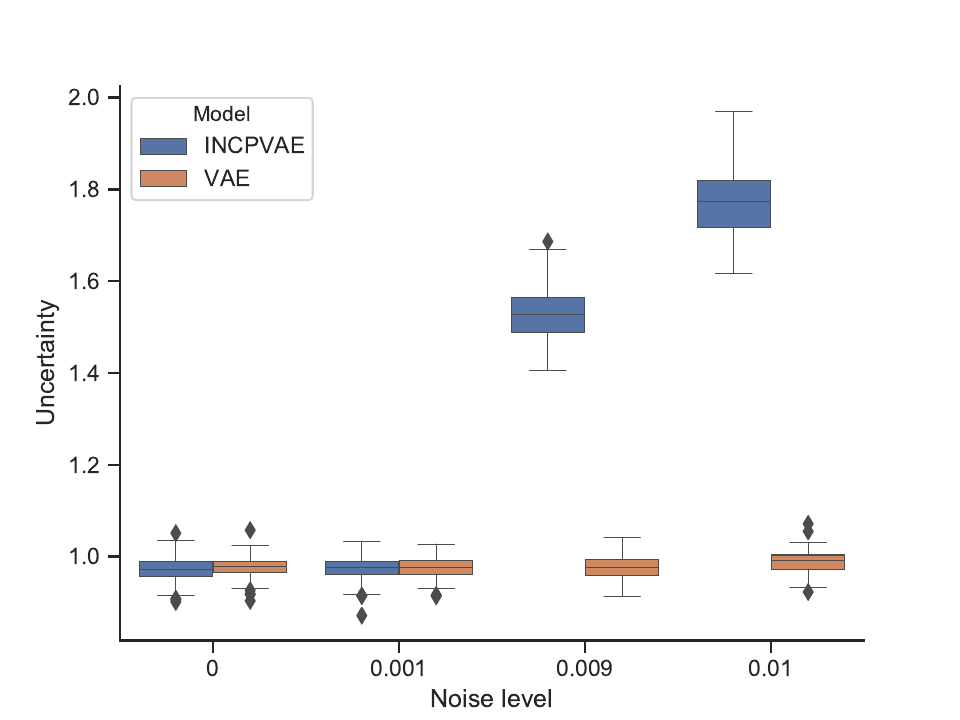}}
\caption[]{Results of the uncertainty estimation task 1. The estimated uncertainty (ELBO ratio, $\mathcal{U}(x)$) from the INCPVAE and traditional VAE model on (a) FashionMNIST, (b) MNIST, (c) CIFAR10, (d) SVHN dataset are presented. Four levels of noise are tested.}
\label{figuncertrain}
\end{figure}

In the uncertainty estimation task 1, we set the standard deviation as the noise level to control the deviation of OOD data from the original data distribution.
We run experiments on FashionMNIST, MNIST, CIFAR10, SVHN datasets, respectively. 
VAE and INCPVAE are trained with the data from the training sets, and then run the inference process with the test samples (OOD data with four levels of noise). The test samples are unseen by models during training process. We calculate the ELBO ratio of the traditional VAE and INCPVAE to estimate the uncertainty on these four datasets. The ELBO ratio is introduced in \textbf{Sec~\ref{incpkl}}. Both the ELBO ratio of traditional VAE and INCPVAE are calculated for 1000 random samples from the testing sets. We then compare the ELBO ratio from INCPVAE and VAE (\textbf{Fig~\ref{figuncertrain}}).

\begin{figure}[H]
\centering
\includegraphics[width=0.95\textwidth]{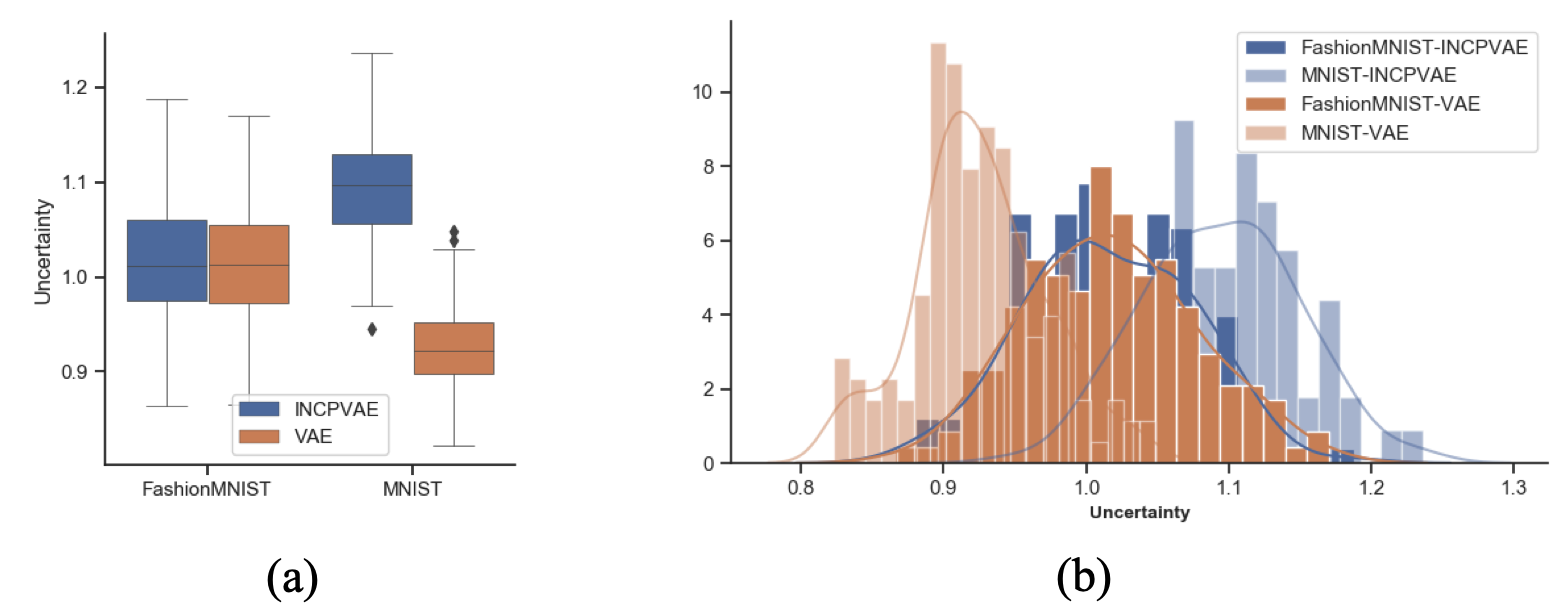}
\caption[]{\re{Results of the uncertainty estimation task 2. The INCPVAE and VAE models are trained on FashionMNIST data, and tested on both FashionMNIST and MNIST. (a) The boxplot of the estimated uncertainty (ELBO ratio, $\mathcal{U}(x)$) from the INCPVAE and traditional VAE model on FashionMNIST and MNIST data. (b) The histogram of the estimated uncertainty from the INCPVAE and traditional VAE model on FashionMNIST and MNIST.}}
\label{figuncerfahion}
\end{figure}

\re{In the uncertainty estimation task 2, we train the model on FashionMNIST dataset, and compare the uncertainty estimation of INCPVAE and standard VAE on FashionMNIST (as ID data) and MNIST (as OOD data). This procedure tests whether the capability of uncertainty estimation in one specific data set can be transferred to another dataset. The results are showed in \textbf{Fig~\ref{figuncerfahion}}.}

In the OOD detection task, we apply the INCP-KL Ratios (defined in \textbf{Sec~\ref{ood}}) as a criterion for OOD detection using INCPVAE model. The tasks are conducted on four pairs of datasets (the training set and the test set).
Specifically, we train INCPVAE and VAE with the samples only from the training set and then compute INCP-KL Ratios of 1000 random samples from the OOD test set. More details about the settings for OOD detection task are in \textbf{Appendix B}. 
We quantify the performance of OOD detection task with INCP-KL Ratios (See results in \textbf{Fig~\ref{figLikelihood}}). 
Moreover, we compare INCPVAE method with 7 existing OOD detection methods, including two likelihood ratio methods, ONID, Mahalanobis distance method, Ensemble method, and WAIC method. We compare INCP-KL of INCPVAE and likelihood of VAE, as well as other baseline methods. The area under the ROC curve (AUROC) and the area under the precision-recall curve (AUPRC) are used as metrics for performance evaluation (See results in \textbf{Table \ref{sample-table0}} and \textbf{Table \ref{sample-table1}}).

More details related to the network architecture and implementation are shown in \textbf{Appendix C}. The code will be available at GitHub.

\subsection{Results of Uncertainty Estimation}
\label{uer}


From \textbf{Fig~\ref{figuncertrain}}, we obtained reliable patterns from these four datasets. When the testing data is drawn without additional perturbations (the noise level is 0), INCPVAE and VAE model present similar uncertainty, suggesting that our model is consistent with standard VAE when it is applied to the ID data. As the noise level increases from 0.01 to 0.1, the INCPVAE-estimated uncertainty of the OOD samples gradually increases in all four datasets, whereas the VAE-estimated uncertainty only shows a slight increase in FashionMNIST dataset and maintains unchanged in the other 3 datasets (MNIST, CIFAR10 and SVHN). These results demonstrate that our INCPVAE model has a strong capability of capturing substantial peculiarity of ID and OOD data with outstanding robustness. Futhermore, we illustrate the ELBO$_I(x)$ from VAE and INCPVAE in~\textbf{Fig~\ref{fig-A1}}, where the standard VAE and our INCPVAE were trained and no noise were imposed during testing. Interestingly, we found that INCPVAE and VAE present almost coincident likelihood distributions in these four datasets, implying that INCPVAE model can reserves the generative ability of VAE model. 

\re{Moreover, \textbf{Fig~\ref{figuncerfahion}}) showed the estimated uncertainty of the test samples from FashionMNIST and MNIST dataset. The INCPVAE obtains higher uncertainty for the OOD data (MNIST) than the ID data (FashionMNIST) during the test step, suggesting that uncertainty estimation of INCPVAE trained in FashionMNIST can be successfully transferred to MNIST dataset. In contrast, VAE showed an opposite trend, which is contradictory to the reality. }

\subsection{Results of OOD Detection}
\label{ood}
We firstly conduct the OOD detection experiments on FashionMNIST and CIFAR10 datasets using a standard VAE model.  \textbf{Fig~\ref{fig-A2}} depicts that the VAE model assigns the OOD data higher likelihoods than training ID data, replicating the nerve-wracking and tricky OOD problem in the likelihood models. 

\begin{figure}[H]
\centering
\includegraphics[width=0.90\textwidth]{{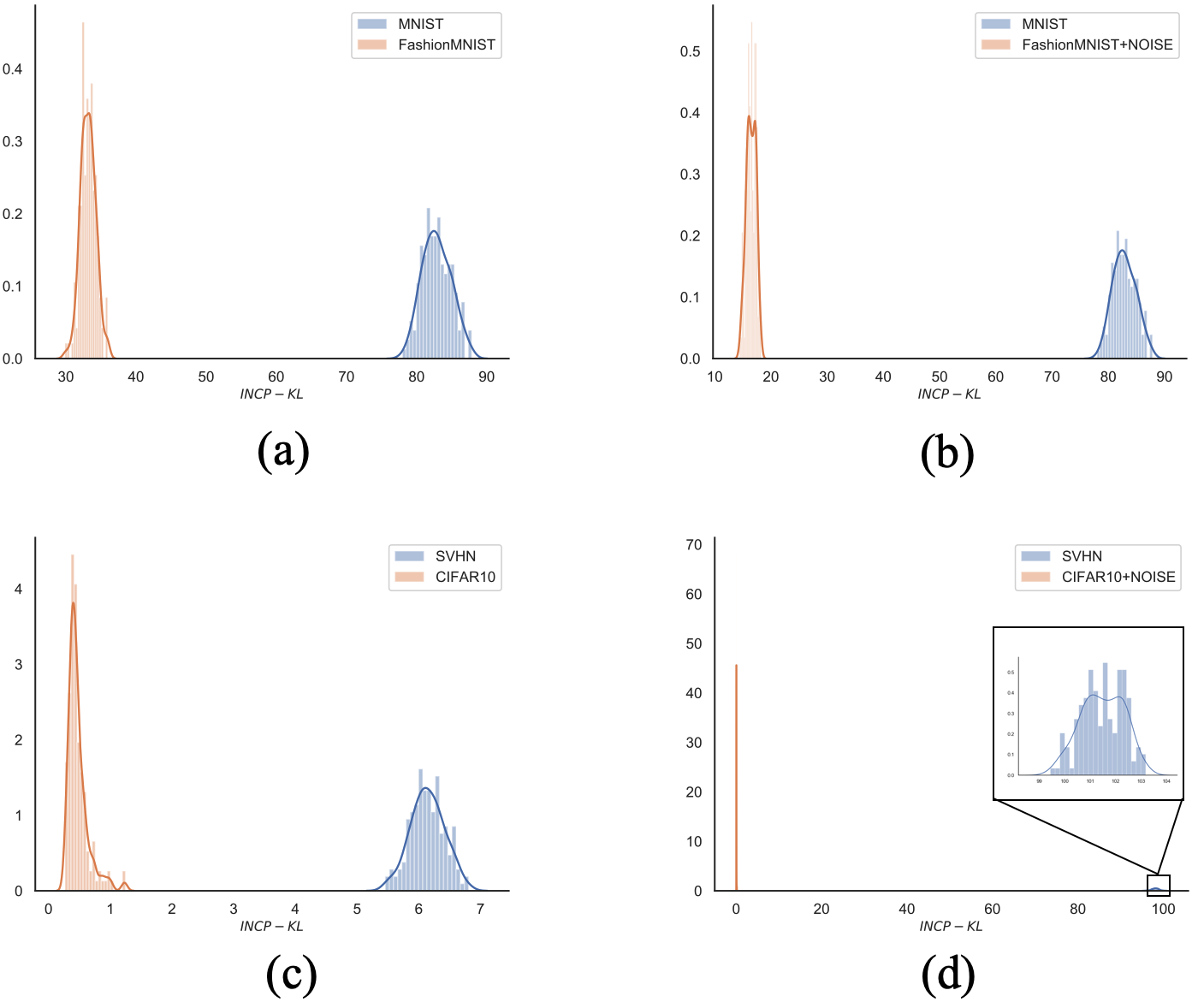}}
\caption[]{\re{The INCP-KL ratio of the INCPVAE in the OOD Detection task. The INCPVAE model is (a) trained on FashionMNIST (ID training set) and FashionMNIST+Noise (OOD training set), and tested on FashionMNIST (ID test set) and MNIST (OOD test set); (b) trained on FashionMNIST+Noise (ID) and FashionMNIST (OOD), and tested on FashionMNIST+Noise (ID) and MNIST (OOD); (c) trained on CIFAR10 (ID) and CIFAR10+Noise (OOD), and tested on CIFAR10 (ID) and SVHN (OOD); (d) trained on CIFAR10+Noise (ID) and CIFAR10 (OOD), and tested on CIFAR10+Noise (ID) and SVHN (OOD). The orange lines are the INCP-KL ratios for ID test data, and the blue lines are for OOD test data. Our results show that the INCP-KL ratios of INCPVAE can largely separate ID and OOD inputs. } }
\label{figLikelihood}
\end{figure}

\textbf{Fig~\ref{figLikelihood}} show the INCP-KL ratio in 4 OOD detection tests using INCPVAE model. \re{Specifically, in OOD test 1 \textbf{Fig~\ref{figLikelihood}(a)}, INCPVAE is trained on FashionMNIST (as ID training set) and FashionMNIST plus noise (as OOD training set), and then test on FashionMNIST (ID test set) and MNIST (OOD test set). 
In OOD test 2 (\textbf{Fig~\ref{figLikelihood}(b)}), INCPVAE is trained on FashionMNIST plus noise (as ID training set) and FashionMNIST (as OOD training set), and then test on FashionMNIST (ID test set) and MNIST (OOD test set). 
In OOD test 3 (\textbf{Fig~\ref{figLikelihood}(c)}), INCPVAE is trained on CIFAR10 (as ID training set) and CIFAR10 plus noise (as OOD training set), and then test on CIFAR10 (ID test set) and SVHN (OOD test set). 
In OOD test 4 (\textbf{Fig~\ref{figLikelihood}(d)}), INCPVAE is trained on CIFAR10 plus noise (as ID training set) and CIFAR10 (as OOD training set), and then test on CIFAR10 plus noise (ID test set) and SVHN (OOD test set). 
It is consistent that the OOD data have higher INCP-KL ratios than the ID data. Together, these results indicates that  INCP-KL ratios for the ID test set and the OOD test set have no overlaps, thus a simple threshold on INCP-KL ratios can detect the OOD data. }

To comprehensively compare the OOD detection performance of different methods, we perform the OOD detection task using INCPVAE and a variety of baseline models. \textbf{Table \ref{sample-table0}} and \textbf{Table \ref{sample-table1}} list the AUROC and AUPRC metrics on the OOD detect tasks (FashionMNIST vs. MNIST, and CIFAR10 vs. SVHN, respectively). Evidently, our model achieves the highest AUROC and AUPRC scores on both tests, compared with other baseline methods.


\begin{table}
  \caption{AUROC and AUPRC for detecting OOD inputs using our INCP-KL Ratio method, likelihood method and other baseline methods on FashionMNIST vs. MNIST datasets.}
  \label{sample-table0}
  \centering
  \begin{tabular}{lll}
    \toprule
    Model     & AUROC    & AUPRC  \\
    \midrule
    \re{INCP-KL} Ratio(Baseline+Noise)  & $\bm{1.000}$  & $\bm{1.000}$  \\
    \re{INCP-KL} Ratio(Baseline)  & $\bm{1.000}$  & $\bm{1.000}$ \\
    Likelihood \re{(Traditional VAE)} & 0.035  & 0.313    \\
    Likelihood Ratio($\mu$)~\cite{ren2019likelihood}     & 0.973 & 0.951      \\
    Likelihood Ratio($\mu$, $\lambda$)~\cite{ren2019likelihood}    & 0.994 & 0.993      \\
    ODIN ~\cite{liang2017enhancing}    & 0.752       & 0.763  \\
    Mahalanobis distance~\cite{lee2018simple} & 0.942  & 0.928  \\
    Ensemble, 20 classifiers~\cite{lakshminarayanan2017simple} & 0.857 & 0.849  \\
    WAIC, 5 models ~\cite{choi2018waic} & 0.221  & 0.401  \\

    \bottomrule
  \end{tabular}
\end{table}

\section{Discussion and Conclusion}

\re{In this study, we have proposed a novel VAE model, called INCPVAE, for reliable uncertainty estimation and OOD detection. Specifically, we firstly improve the noise contrastive prior, called INCP, to be suitable for VAE models, and then present a hybrid method combining INCP with the encoder of VAE framework. 
Using INCPVAE model, OOD samples can be generated by adding Gussian noise into the ID samples; therefore, INCPVAE model can be jointly trained with ID data and OOD data. We define a new metric (ELBO Ratio) for uncertainty estimation and a new OOD detection criterion which is based on INCP-KL Ratio. }


We reproduced the results that traditional VAE easily assigns higher likelihoods for OOD samples than ID samples (\textbf{Fig~\ref{fig-A2}}). \re{These results suggest that the likelihood in traditional VAEs is not a good metric to detect the OOD data, which is consistent with previous studies~\cite{nalisnick2018deep, hendrycks2018deep,choi2018waic,lee2017training,nalisnick2019detecting} and the model with reliable uncertainty estimation can improve the performance of OOD detection.}
\re{Firstly, 
we proposed a new metric, ELBO Ratio. The result of the uncertainty estimation task 1 across four datasets (\textbf{Fig~\ref{figuncertrain}}) demonstrates that ELBO Ratio increases as the noise increases. The uncertainty estimation task 2 shows that INCPVAE trained with FashionMNIST data can accurately estimate the uncertainty in MNIST data, whereas the VAE model failed to transfer the uncertainty information (\textbf{Fig~\ref{figuncerfahion}}). Together, these results indicate that ELBO Ratio can reliably index the uncertainty in the input data.}

\begin{table}
  \caption{AUROC and AUPRC for detecting OOD inputs using INCP-KL Ratio method, likelihood method, and other baselines on CIFAR10 vs. SVHN datasets.}
  \label{sample-table1}
  \centering
  \begin{tabular}{lll}
    \toprule
    Model     & AUROC    & AUPRC  \\
    \midrule
    \re{INCP-KL} Ratio(Baseline+Noise)  & $\bm{1.000}$  & $\bm{1.000}$  \\
    \re{INCP-KL} Ratio(Baseline)  & $\bm{1.000}$  & $\bm{1.000}$ \\
    Likelihood \re{(Traditional VAE)} & 0.057 & 0.314     \\
    Likelihood Ratio($\mu$)~\cite{ren2019likelihood}    &  0.931  & 0.888      \\
    Likelihood Ratio($\mu$, $\lambda$)~\cite{ren2019likelihood}     & 0.930 & 0.881  \\
    \bottomrule
  \end{tabular}
\end{table}

\re{Secondly, we proposed a metric called INCP-KL ratio to detect OOD data. A simple threshold on INCP-KL ratios (e.g. $\alpha = 1$ in Eq.\ref{ood-detection}) can be used to detect OOD data in INCPVAE model.} The results of OOD detection task demonstrate that our model achieves SOTA performance to differentiate OOD and ID data, compared with baseline methods (\textbf{Table \ref{sample-table0}} and \textbf{Table \ref{sample-table1}}). INCPVAE model, as a model-independent method to OOD detection, paves a way for future VAE applications on OOD detection. Also, INCPVAE can be easily extended to anomaly detection and adversarial example detection.

\re{Despite the advantages of our work, there are still some limitations and future work worth mentioning. We only focused on the uncertainty estimation and OOD detection using VAE model in this study. It is interesting to extend INCP to other generative models, such as GAN. Moreover, we generate OOD data by adding Gaussian noise to ID data, which cannot capture the characteristics of the OOD data in the real applications. Other methods to generate appropriate OOD inputs are worthy of investigation; for example, using GAN to generate OOD data~\cite{lee2017training}. The realistic OOD data can help to train INCPVAE models, as it can be potentially used to generate priors of INCPVAE. Alternatively, adversarial examples~\cite{goodfellow2014explaining} may also be used to train INCPVAE, in order to enhance robustness of VAE. }

\re{In summary, we integrate INCP into VAE framework to solve the problem that the OOD detection techniques for deep generative models are hardly transferred to VAEs~\cite{xiao2020likelihood}.}

\section*{Acknowledgements}
The authors thank the anonymous reviewers for their comment sand Dr.Steffen Bollmann for his suggestion.This work was funded in part by the National Natural Science Foundation of China (62001205), Guangdon g Natural Science Foundation Joint Fund (2019A1515111038), Shenzhen Science and Technology Innovation Committee (20200925155957004, SGDX2020110309280
100, KCXFZ2020122117340001), Shenzhen Key Laboratory of Smart Healthcare Engineering (ZDSYS20200811144003009), CAAI-Huawei Mindspore Open Fund (CAAIXSJLJJ-2020-024A), Fundamental Research Funds for Central Universities(DUT21RC(3)091), Beijing Science and Technology Programs (Z191100007 519009).


\newpage
\appendix
\section{Settings for uncertainty estimation}

In this section, we introduce detailed settings for uncertainty estimation. To evaluate uncertainty estimation from the traditional Variational Auto-encoder (VAE) and from the Improved Noise Contrastive Priors VAE (INCPVAE), we train VAE on in-distribution (ID) training set and INCPVAE on the ID and out-of-distribution (OOD) training set. Then we test both of VAE and INCPVAE on ID testing set and OOD testing set0/set1/set2, respectively. See full lists in \textbf{Table~\ref{table1}}. The OOD training set and testing set0/set1/set2 are generated by adding three levels of Gaussian noise to the baseline (See~\textbf{Table~\ref{table2}}).

\begin{table}[H]
\centering
  \caption{Baselines are FashionMNIST, MNIST, CIFAR10,SVHN.
  Noise is generated by Gussian Noise($\mu,\sigma^2$), where $\mu=0$, $\sigma=\sigma_{0},\sigma_{1},\sigma_{2}$.}
  \label{Stable1}
  \centering
  \begin{tabular}{lll}
  \toprule
  \cmidrule(r){1-3}
   Dataset      & VAE        &  INCPVAE    \\
  \midrule
    ID training set       & Baseline & Baselline  \\
    OOD training set  & - &  Baseline+Noise($\sigma_{1}$)\\ \hline
    ID tesing set          & Baseline &Baseline     \\
    OOD tesing set0     & Baseline+Noise($\sigma_{0}$)   & Baseline+Noise($\sigma_{0}$)  \\
    OOD tesing set1   &Baseline+Noise($\sigma_{1}$) &Baseline+Noise($\sigma_{1}$)\\
    OOD tesing set2   &Baseline+Noise($\sigma_{2}$) &Baseline+Noise($\sigma_{2}$)  \\ 
    \bottomrule
  \end{tabular}
  \label{table1}
\end{table}

\begin{table}[H]

 \caption{The levels of noises added to four baseline datasets. Noise is generated by Gussian Noise($\mu,\sigma^2$), where $\mu=0$, $\sigma=\sigma_{0},\sigma_{1},\sigma_{2}$.}
 \centering
 \begin{tabular}{ccccc}
 \toprule
 \cmidrule(r){1-5}
   Noise level     & FashionMNIST     &  MNIST  & CIFAR10 & SVHN    \\
 \midrule
  $\sigma_{0}$       & 0.0001 & 0.001& 0.01 &0.001 \\
    $\sigma_{1}$       & 0.00028 & 0.008 & 0.05 &0.009 \\
    
     $\sigma_{2}$       & 0.1000 & 0.010 & 0.10 &0.010 \\
\bottomrule
\end{tabular}
\label{table2}
\end{table}

\begin{table}[H]
 \caption{ True OOD posterior of INCPVAE ${\tilde p}(\bm \tilde z \mid \bm \tilde x)$ is employed by Gaussian distribution $\mathcal{N}(\mu_{ \bm \tilde x },\sigma^2_{\bm \tilde x})$.}

 \centering
 \begin{tabular}{lc}
 \toprule
 \cmidrule(r){1-2}
   Dataset & Uncertainty level ($\sigma_{\bm \tilde{x}}$) \cr
  \midrule
  FashionMNIST & $e^{0.65}$ \cr
  MNIST & $e^{0.65}$ \cr
  CIFAR10 & $e^{1.00}$ \cr
  SVHN & $e^{1.00}$ \cr
    \bottomrule
  \end{tabular}
   \label{table3}
\end{table}

For each image dataset, the true OOD posterior of INCPVAE (or OOD data output prior) is assumed by Gaussian distribution with a specific variance (See \textbf{Table~\ref{table3}}), which represents that these four datasets have various uncertainties.

\begin{figure}[H]
\centering
\subfigure[FashionMNIST]{\includegraphics[width=0.4\textwidth]{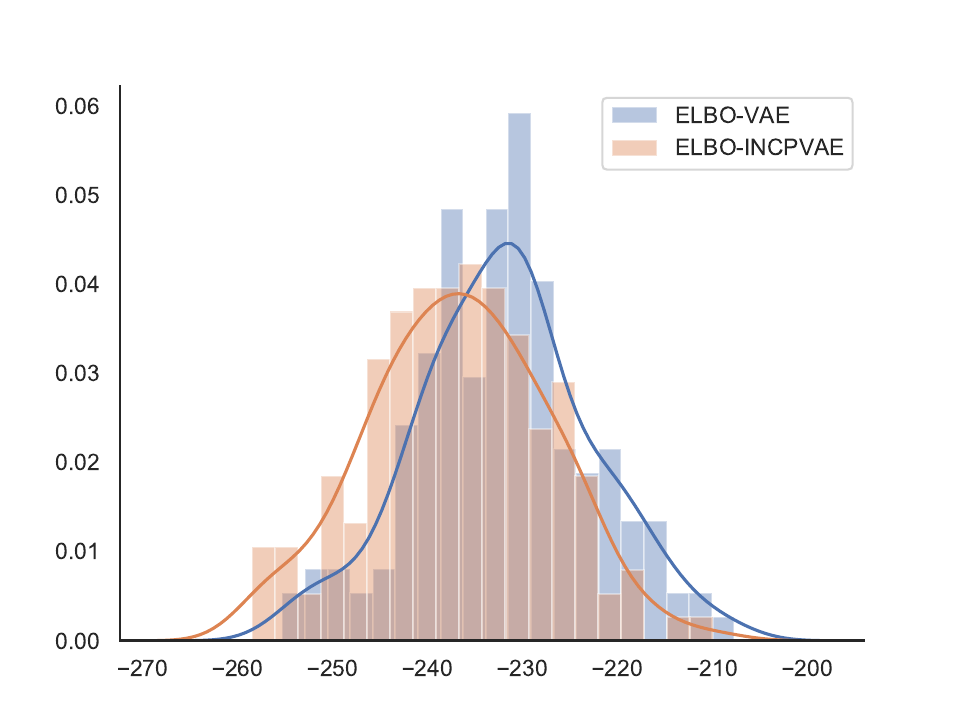}}
\subfigure[MNIST]{\includegraphics[width=0.4\textwidth]{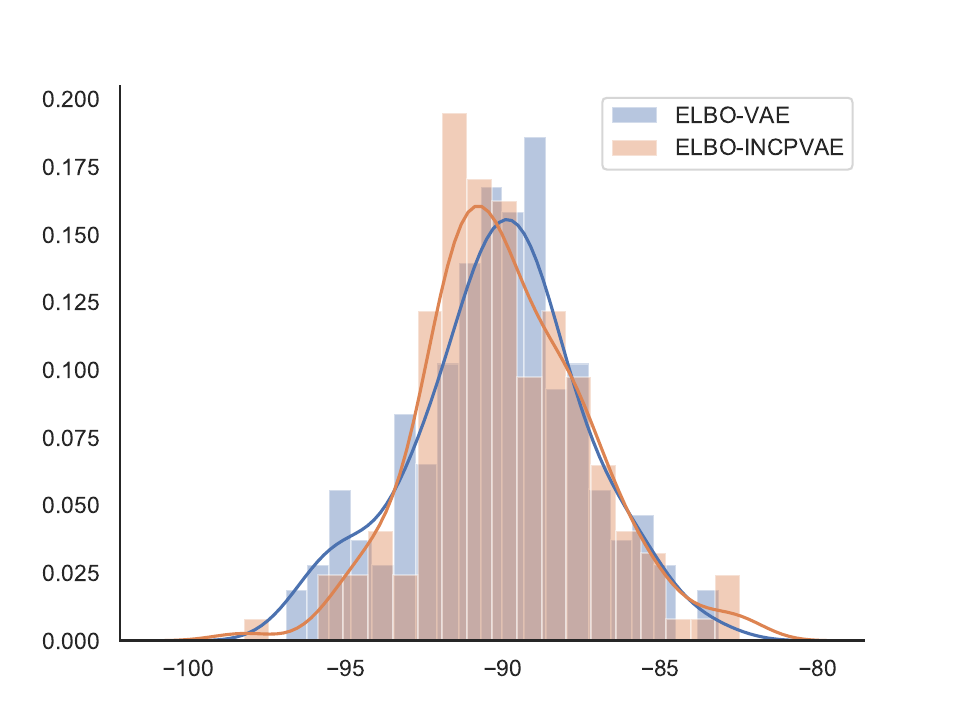}} 
\subfigure[CIFAR10]{\includegraphics[width=0.4\textwidth]{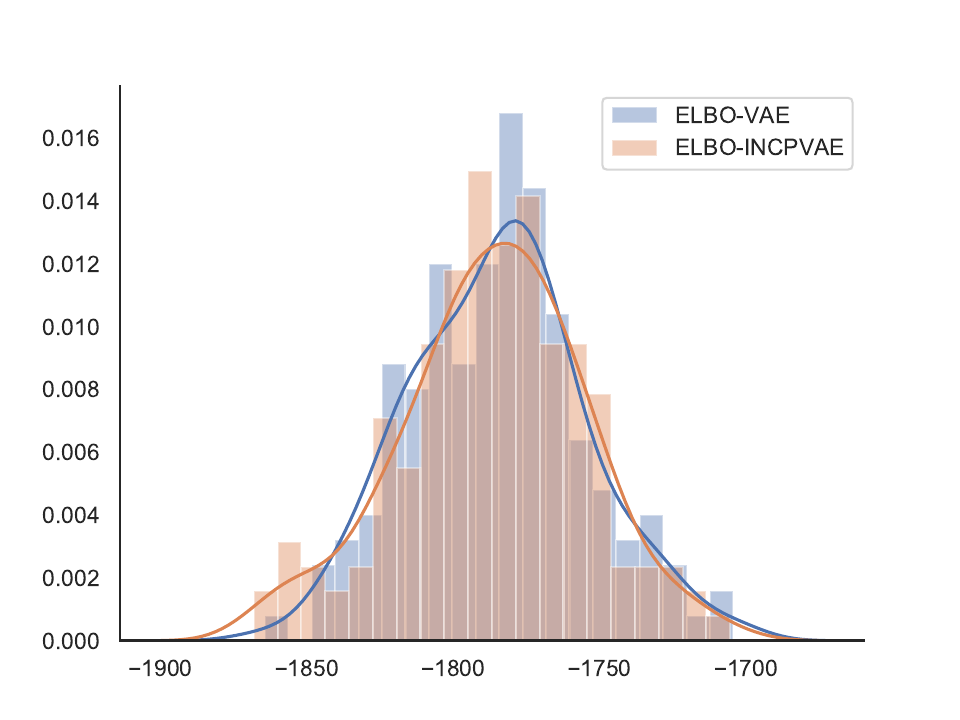}}
\subfigure[SVHN]{\includegraphics[width=0.4\textwidth]{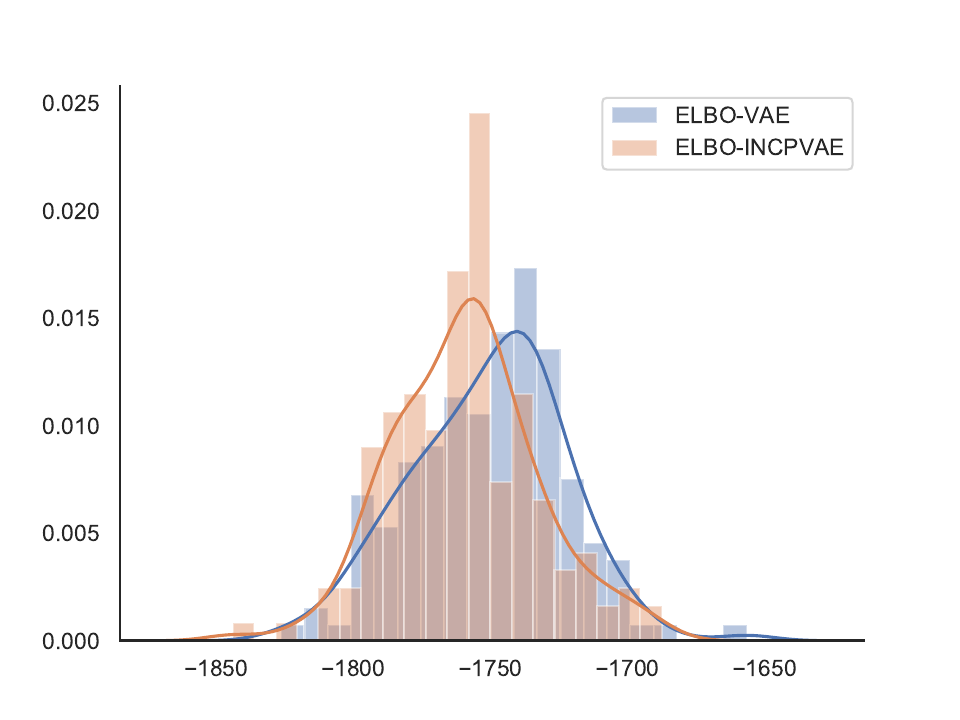}} 
\caption[]{The histogram of the ELBO of the ID data, ELBO$_I(x)$, for VAE and INCPVAE. (a) FashionMNIST, (b) MNIST, (c) CIFAR10, (d) SVHN dataset. These results demonstrate that INCPVAE has similar ELBO$_I(x)$ with VAE.}
\label{fig-A1}
\end{figure}

\section{Settings for OOD Detection}

In this section, we introduce detailed settings of OOD detection experiments. Firstly, following the most challenging experiment reported by Nalisnick {\textit{et al.}}, we train VAE on ID training set and test on ID and OOD testing set (See \textbf{Table~\ref{table4}}).
Secondly, to evaluate the OOD detection of INCPVAE, we train INCPVAE on the ID and OOD training set, and test INCPVAE on OOD testing set and OOD testing set1 (See \textbf{Table~\ref{table5}}). The ID and OOD training set, as well as the OOD testing set, are generated by adding Gaussian noise with three levels to baseline(See~\textbf{Table~\ref{table6}}).

\begin{table}[H]
 \caption{ Datasets: VAE for OOD detection }
 \centering
 \begin{tabular}{l|c|cc}
  \hline
   Exp & ID training set  & ID test set  & OOD test set \\ 
  \hline
   Exp1 & FashionMNIST & FashionMNIST & MNIST \\
   Exp2 & CIFAR10 &  CIFAR10 & SVHN    \\
  \hline
  \end{tabular}
   \label{table4}
\end{table}

\begin{table}[H]
\caption{Datasets for INCP-KL Ratios of INCPVAE. Fashion is short for FashionMNIST.}
\centering
\begin{adjustbox}{width=1\columnwidth}
\begin{tabular}{c|cc|cc}
\hline
Exp   & ID training set & OOD training set & OOD test set1 & OOD test set2  \\
\hline
Exp1 & Fashion & Fashion+Noise($\sigma_3$) & Fashion+Noise($\sigma_3$) & MNIST \\
Exp2 & Fashion+Noise($\sigma_4$) & Fashion & Fashion & MNIST \\
Exp3 & CIFAR10 & CIFAR10+Noise($\sigma_3$) & CIFAR10+Noise($\sigma_3$) & SVHN \\
Exp4 & CIFAR10+Noise($\sigma_4$) & CIFAR10 & CIFAR10 & SVHN \\
\hline
\end{tabular}
\end{adjustbox}
\label{table5}
\end{table}

\begin{table}[h]

  \caption{Datasets for INCP-KL Ratios of INCPVAE. Noise is generated by Gussian Noise($\mu,\sigma^2$), where set $\mu=0$, $\sigma=\sigma_{3},\sigma_{4}$}

  \centering
  \begin{tabular}{lcc}
  \toprule
  \cmidrule(r){1-3}
   dataset    &   Noise level ($\sigma_{3}$) &   Noise level ($\sigma_{4}$)  \\
  \midrule
    FashionMNIST    & 0.00028   & 0.00050  \\
    CIFAR10         & 0.05000   & 0.09000  \\
    \bottomrule
  \end{tabular}
  \label{table6}
\end{table}

For different datasets, the true OOD posterior of INCPVAE (or OOD data output prior) is Gaussian distribution with different variance (See \textbf{Table~\ref{tables}}), which represents that different datasets have different uncertainties.

\begin{table}[H]
 \caption{ True OOD posterior of INCPVAE ${\tilde p}(\bm \tilde z \mid \bm \tilde x)$ is employed by Gaussian distribution $\mathcal{N}(\mu_{ \bm \tilde x },\sigma^2_{\bm \tilde x})$.}

 \centering
 \begin{tabular}{lc}
 \hline
   Dataset & Uncertainty level ($\sigma_{\bm \tilde{x}} $)       \cr
 \hline
  FashionMNIST & $e^{0.65}$ \cr
  CIFAR10 & $e^{1.00}$ \cr
  \hline
  \end{tabular}
   \label{tables}
\end{table}

\begin{table}[H]
\centering
\vskip -0.2cm
\caption{Encoder architecture. This architecture was used for VAE and INCPVAE trained on FashionMNIST with linear layer units 3136 and CIFAR10 with 4096.}
\begin{tabular}{ccccc}
\hline
\textbf{Operation} & \textbf{kernel} & \textbf{stride} & \textbf{Features} &  \textbf{padding} \\ \hline
Input & - & - & - & - \\
Convolution & $5 \times 5$  & $2 \times 2$  & 256    & 0    \\
Convolution & $5 \times 5$  & $2 \times 2$  & 32     & 0    \\
Convolution & $5 \times 5$  & $1 \times 1$  & 32     & 0    \\ 
Dense      & -   & - & 3136/4096   & -  \\
\hline
\end{tabular}
\label{Encoder}
\end{table}

\begin{figure}[H]
\centering
\subfigure[VAE]{\includegraphics[width=0.45\textwidth]{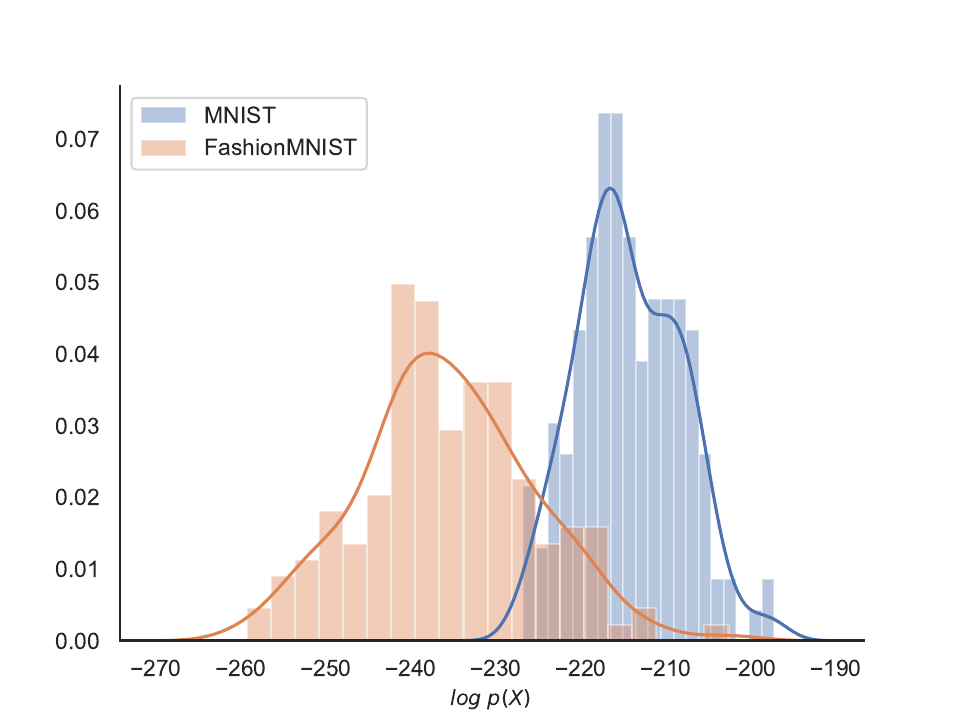}}
\subfigure[VAE]{\includegraphics[width=0.45\textwidth]{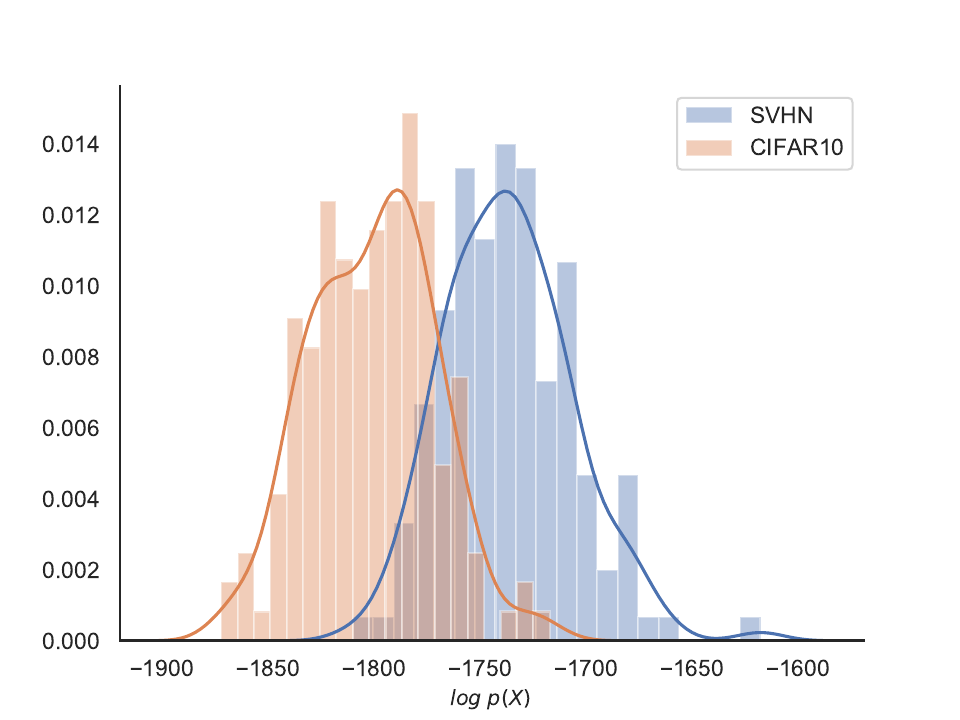}}
\caption[]{\re{The histogram of the marginal likelihood of the VAE. (a) VAE trained on FashionMNIST (ID), and tested on FashionMNIST and MNIST (OOD); (b) VAE trained on CIFAR10 (ID), and tested on CIFRAR10 (ID) and SVHN (OOD). The orange lines are for ID data, and the blue lines are for OOD data.}}
\label{fig-A2}
\end{figure}

\section{Settings for Implementation Detail}

In the experiments, VAE and INCPVAE are trained on FashionMNIST and CIFAR10. All models are trained with images normalized to $[0,1]$ on 1 $\times$ NVIDIA TITAN RTX GPU. In all experiments, VAE and INCPVAE consist of an encoder with the architecture given in \textbf{Table~\ref{Encoder}} and a decoder shown in \textbf{Table~\ref{Decoder}}. Both VAE and INCPVAE use Leaky Relu activation function. We train the VAE for 200 epochs with a constant learning rate $1e^{-4}$, meanwhile using Adam optimizer and batch size 64 in each experiment.

\begin{table}[H]
\centering
\caption{Decoder architecture. This architecture was used for VAE and INCPVAE trained on FashionMNIST with linear layer units 3136 and CIFAR10 with 4096.}
\begin{tabular}{ccccc}
\hline
\textbf{Operation} & \textbf{kernel} & \textbf{stride} & \textbf{Features} & \textbf{padding} \\ \hline
Input $\bm{z}$ & - &- &- &-\\
Dense  & -    & -   & 3136/4096 & -   \\
Dense  & -    & -   & 1568/2048 & -  \\
Transposed Convolution & $5 \times 5$ & $1 \times 1$ & 32   & 0      \\
Transposed Convolution & $5 \times 5$ & $2 \times 2$ & 256  & 0                \\ 
Transposed Convolution & $5 \times 5$ & $2 \times 2$ & 3  & 0                \\ 
\hline
\end{tabular}

\label{Decoder} 
\end{table}



\end{document}